\pdfoutput=1

\documentclass[11pt]{article}

\usepackage[final]{acl}

\usepackage{times}
\usepackage{latexsym}

\usepackage[T1]{fontenc}

\usepackage[utf8]{inputenc}

\usepackage{microtype}

\usepackage{inconsolata}

\usepackage{CJKutf8}
\newcommand{\cjk}[1]{\begin{CJK}{UTF8}{ipxm}{#1}\end{CJK}}
\usepackage{graphicx}
\usepackage{enumitem}

\usepackage{array}
\usepackage{longtable}
\usepackage{colortab}
\usepackage{colortbl}
\usepackage{arydshln}
\usepackage{multirow}
\usepackage{amsmath} 
\renewcommand{\u}{\underline}

\title{Knowledge of Pretrained Language Models \\on Surface Information of Tokens}

\author{
  \textbf{Tatsuya Hiraoka}$^{\dagger\ddagger}$\qquad
  \textbf{Naoaki Okazaki}$^\dagger$
  \\\\
  $^\dagger$ Tokyo Institute of Technology \qquad $^\ddagger$ Fujitsu Limited\\
  \texttt{tathi029@gmail.com} \quad \texttt{okazaki@c.titech.ac.jp}
  }

\begin{document}
\maketitle
\begin{abstract}
Do pretrained language models have knowledge regarding the surface information of tokens?
We examined the surface information stored in word or subword embeddings acquired by pretrained language models from the perspectives of token length, substrings, and token constitution.
Additionally, we evaluated the ability of models to generate knowledge regarding token surfaces.
We focused on 12 pretrained language models that were mainly trained on English and Japanese corpora.
Experimental results demonstrate that pretrained language models have knowledge regarding token length and substrings but not token constitution.
Additionally, the results imply that there is a bottleneck on the decoder side in terms of effectively utilizing acquired knowledge.
\end{abstract}

\section{Introduction}

Pretrained language models (PLMs) are widely used for various natural language processing (NLP) tasks~\cite{peters2018deep,radford2018improving,devlin2019bert}.
In particular, pretrained large language models (LLMs) contribute to considerable performance improvements on generation tasks~\cite{brown2020language,chen2021evaluating,chang2023survey,min2023recent,zhao2023survey}.

\begin{figure}[t]
    \centering
    \includegraphics[width=7.0cm]{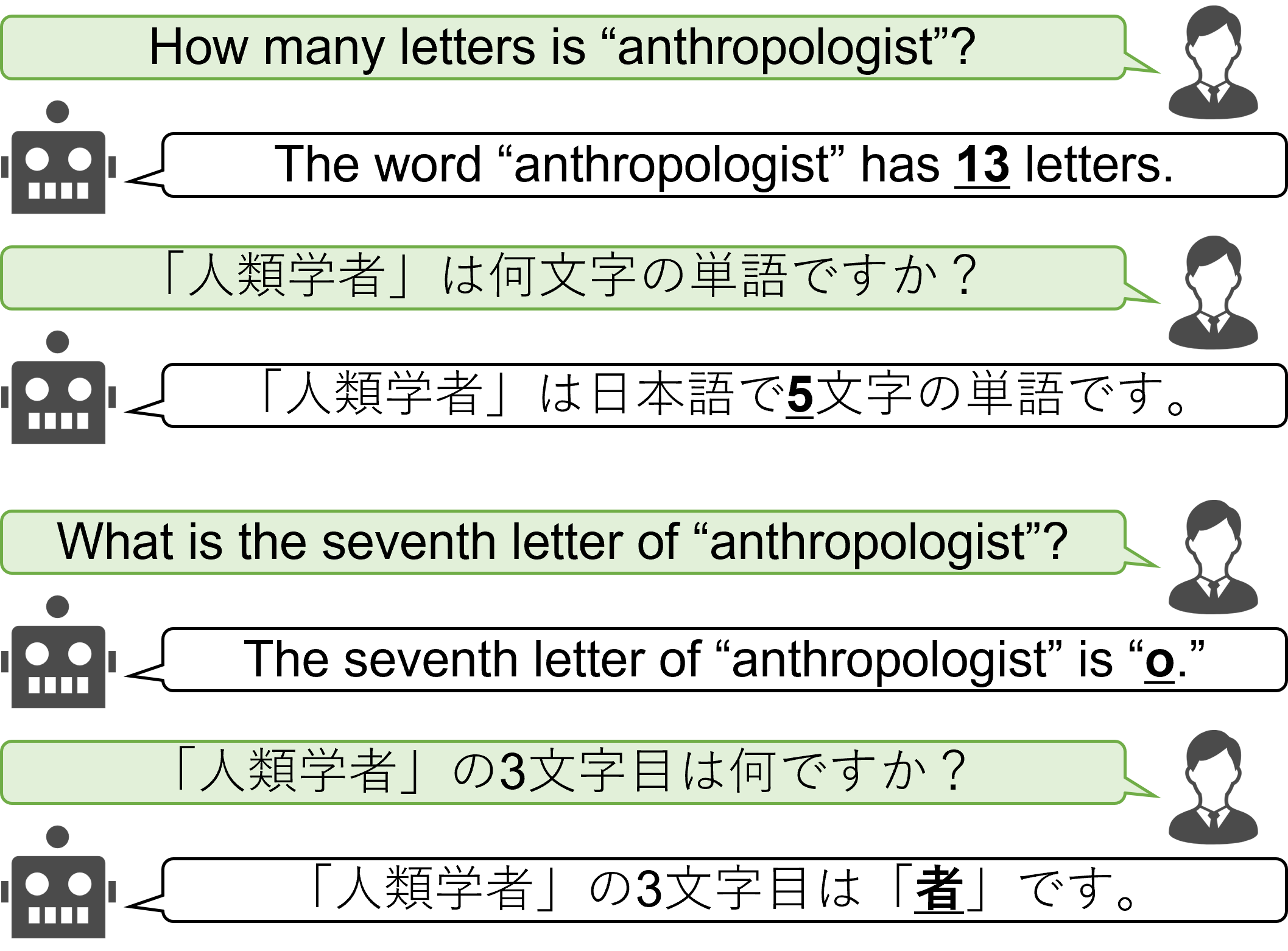}
    \caption{
        Input and output examples when asking GPT-3.5 Turbo about the surface information of words (as of 1st, Jan. 2024).
        The Japanese example has the same meaning as the English text, asking the length of and third character in \cjk{人類学者} (anthropologist).
    }
    \label{fgr:intro_example}
\end{figure}

Although PLMs are widely used in NLP, their knowledge regarding surface information is still open to discussion.
Current PLMs have difficulty on tasks that require textual surface information such as counting the number of characters, extracting substrings from texts, and generating palindromes~\cite{huang2023inducing}.
For example, when asking for the length of a word and the $N$-th character in the word to the well-known PLM ChatGPT (GPT-3.5 Turbo), incorrect answers are obtained for some long words in English (Figure \ref{fgr:intro_example}).
Furthermore, when asking the same question for non-English texts such as Japanese texts, it is difficult to obtain the correct answer even for short words. 
We believe that the lack of surface knowledge of PLMs is a significant bottleneck for certain NLP tasks that require models to understand surface information, including information extraction~\cite{ma2022named,wang2023gpt}, length-specified generation~\cite{fan2018controllable,takase2019positional,dufter2022position}, and extractive question answering~\cite{yoon2019pre,jiang2021can}.

This problem may be a result of the typical  architectures of recent PLMs, whose inputs are subword-level embeddings that cannot directly access character-level surface information.
Some architectures with character-level inputs appear to improve performance on NLP tasks~\cite{sun2023characters}.
Although a few studies have examined the knowledge of PLMs regarding surface information~\cite{kaushal2022tokens,huang2023inducing}, it is uncertain which types of surface information are stored in the subword- and word-level embeddings learned by PLMs.
We believe that the first step in developing \textit{surface-sensitive} PLMs is to analyze the knowledge of current PLMs regarding surface information.

In this study, we examined whether the embeddings of PLMs contain knowledge regarding the surface information of texts from three perspectives: A) information regarding token length, B) information regarding substrings, and C) information regarding token constitution.
According to the results of experiments conducted on PLMs trained on English and Japanese corpora, the embeddings of subwords and words contain partial knowledge regarding token length and substrings but no  knowledge regarding token constitution (i.e., which characters are located in which positions).
Additionally, we examined decoder knowledge regarding surface information, specifically word length, and identified a decoder-side bottleneck in terms of utilizing obtained surface information.

\begin{figure*}
    \centering
    \includegraphics[width=15.5cm]{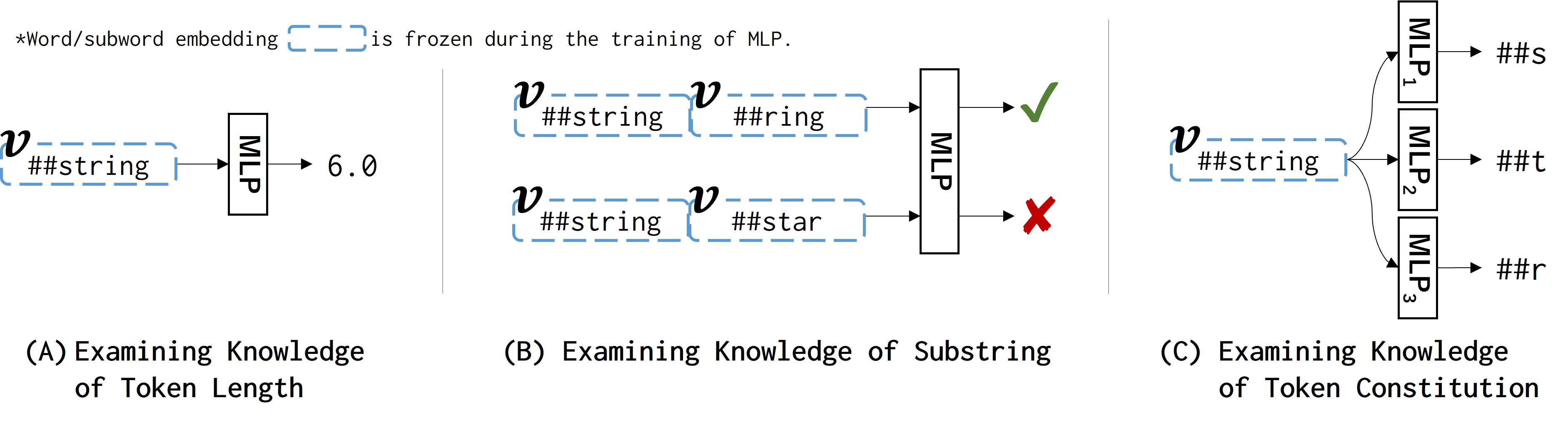}
    \caption{
        Outline of the methods used to examine the obtained knowledge regarding surface information in word/subword embeddings ($\mathbf{v}$ in the figure).
        This figure shows the inputs and outputs of each method.
        We trained independent MLPs for each task.
    }
    \label{fig:outline}
\end{figure*}

\section{Related Work}

Many papers have reported that the language models that consider fine-grained information, such as byte-level~\cite{xue2022byt5} and character-level~\cite{tay2021charformer,clark2022canine} information, or both ~\cite{kim2016character,ma2020charbert,el2020characterbert,sun2023characters}, contribute to performance improvements on various NLP tasks.
This area of research reflects the shortcomings of the surface knowledge acquired by models with subword-level inputs.
If a subword-level model could obtain character information, it would be possible to improve NLP task performance without directly inputting character information, which makes inputs excessively long~\cite{huang2023inducing}.
We believe that our analysis will contribute to the development of PLMs  that can appropriately acquire and utilize surface knowledge.

Tokenization can be a bottleneck in obtaining word-level surface information.
In addition to the discussion of tokenization methods~\cite{sennrich2016neural, kudo2018sentencepiece,song2020linear}, some studies have discussed the tokenization preferences of humans, which are related to surface information~\cite{beinborn2023analyzing,hiraoka2023tokenization}.
In a Transformer architecture~\cite{vaswani2017attention}, positional encoding is important for capturing surface information because many words are encoded as sequences of subwords~\cite{wang2021on,dufter2022position}.

\newcite{kaushal2022tokens} is the most relevant work as it described the knowledge of characters in embeddings.
Our study differs from this work in its terms of its wider scope of analysis.
In addition to the examination of knowledge regarding the characters in texts, which was the focus of the work mentioned above, we present experimental results for knowledge regarding subword-level substrings in texts (see \S \ref{sec:substring-explanation}).
Our experiments included character-level examinations and the results were consistent with those of  previous works.
Furthermore, we present an analysis of surface knowledge regarding token length (see \S \ref{sec:length-explanation}) and token constitution (see \S \ref{sec:character-explanation}) using recently developed LLMs.

\section{General Settings}
\label{sec:outline}
\subsection{Experimental Overview}
We examined the knowledge regarding surface information acquired by PLMs.
In particular, we analyzed the knowledge contained in the subword or word embeddings of PLMs from three perspectives, namely whether the embeddings of PLMs contain information regarding
\vspace{-0.5\baselineskip}
{
\begin{enumerate}[label=(\Alph*)]
    \setlength{\parskip}{0cm}
    \setlength{\itemsep}{0cm}
    \item lengths of subwords/words (\S \ref{sec:length-explanation})
    \item substrings of subwords/words (\S \ref{sec:substring-explanation})
    \item token constitutions of subwords/words (\S \ref{sec:character-explanation})
\end{enumerate}
}
\vspace{-0.5\baselineskip}   
Figure \ref{fig:outline} summarizes the examination methods corresponding to each of these perspectives.

For each examination, we trained an individual multi-layer perceptron (MLP) whose input was the embedding $\mathbf{v}_s$ corresponding to a subword $s \in S$ or embedding $\mathbf{v}_w$ corresponding to a word $w \in W$.
The MLP was trained to each of the examination task.
We analyzed the knowledge regarding surface information stored by PLMs by observing the outputs of the MLPs ($y=f(\mathbf{v}_s)$ or $y=f(\mathbf{v}_w)$) trained on different experimental settings.

In this paper, we report experimental results with average scores over $k$-fold cross-validation, where $k=10$. 
In other words, we used 90\% of the subword/word list ($S$ or $W$)  to train each task-specific MLP and used the remaining 10\% to evaluate each MLP.
We then repeated this training and evaluation procedure for each fold of cross-validation and calculated the average scores.

\subsection{Target PLMs}
\label{sec:target}
To examine the surface knowledge of various types of PLMs, we selected nine pretrained models that were trained mainly on English corpora.
Additionally, we selected three models trained primarily on a non-English Japanese corpus.
We selected Japanese because it has different writing systems for tokenization and a number of different character types compared to English.
Note that it is impossible to compare the results of the models under strictly fair conditions because each model was trained on different corpora with varying vocabulary.
Therefore, we used the experimental results to observe and analyze the major trends of PLMs in terms of capturing surface information.

For conventional embeddings pretrained on English corpora prior to the LLM era, we used \textbf{Word2Vec}~\cite{mikolov2013efficient}, \textbf{FastText}~\cite{bojanowski2016enriching}, and \textbf{GloVe}~\cite{pennington2014glove}\footnote{We refer to classical embedding methods such as Word2Vec as PLMs for the sake of simple explanation.}.
The Word2Vec and FastText models were trained using a skip-gram trainer provided by Gensim~\cite{rehurek2011gensim}.
For both models, we used 20 GB of text data collected from the English version of Wikipedia after the pre-tokenization of punctuation using the English NLTK tokenizer~\cite{bird2006nltk}.
For the GloVe model, we used \verb|glove-wiki-gigaword-300|.

We selected \textbf{BERT-base-cased/uncased} as pretrained masked language models~\cite{devlin2019bert}.
\textbf{T5-base}~\cite{raffel2020t5}, \textbf{LLaMA2-7B}, and \textbf{LLaMA2-7B-chat}~\cite{touvron2023llama} were selected as pretrained single-directional language models.
As a baseline model to validate our examination settings, we also considered \textbf{CANINE}~\cite{clark2022canine}, which is a PLM that uses character-level information as inputs.
We consider CANINE to have an advantage over the other models because character-level information helps complete tasks that require surface information.

We selected \verb|rinna/japanese-gpt2-medium| (\textbf{rinna-medium}) as a GPT-2 architecture~\cite{radford2019language} pretrained on a Japanese corpus\footnote{\url{https://huggingface.co/rinna}}.
Additionally, we considered \verb|Swallow -7b-hf| (\textbf{Swallow-7B}) and \verb|Swallow-7b-instruct-hf| (\textbf{Swallow-7B-instruct})\footnote{\url{https://huggingface.co/tokyotech-llm}}, which are the most recent Japanese PLMs based on LLaMA2.

We did not consider LLMs that utilize external applications such as function calling.
Some LLMs can solve tasks related to surface knowledge (see \S \ref{sec:length-explanation}, \ref{sec:substring-explanation}, \ref{sec:character-explanation}) using a programming module~\cite{balse2023investigating,kiesler2023exploring}. However, we consider such functions to lie outside the capabilities of an LLM itself.
Our interest lies in the knowledge regarding surface acquired in the parameters of PLMs, which can be utilized for downstream tasks through fine-tuning or zero-/few-shot learning.


\subsection{Input Granularity}
This study focused on two input granularities, namely subword-level and word-level inputs.

\subsubsection{Subword level}
\label{sec:gran_subword}
For our examination using subword-level inputs, we simply used the embedding $\mathbf{v}_s$ corresponding to the subword $s$ included in the vocabulary $S$ of the PLMs.
For models with individual embedding parameters for the input and output layers, we used only the input embeddings.
Special tokens such as \verb|<unk>| and tokens for byte fallback were excluded. 

CANINE is out of focus for subword-level examination because all tokens in the CANINE vocabulary are single characters. Therefore, inspecting intra-subword information is meaningless.
We treated the classical word embedding methods (i.e., Word2Vec, FastText, and GloVe) as subword-level embeddings in this study because we used in-vocabulary embeddings in our experiments.

\subsubsection{Word level}
\label{sec:gran_word}
For our examination using word-level inputs, we used a representation $\mathbf{v}_w$ corresponding to the word $w$ included in word list $W$.
We utilized the most frequent 100,000 words in the Project Gutenberg dataset provided by Wiktionary as the word list $W$ for the English models\footnote{We used the word list updated  on August 15, 2005: \url{https://en.wiktionary.org/wiki/Wiktionary:Frequency_lists/English/Project_Gutenberg}. Because of duplicate words, the exact number of tokens was 83,404.}.
For the Japanese models, we used the dictionary of Sudachi~\cite{takaoka2018sudachi}, which is a Japanese morphological analyzer.
We selected 795,336 A-mode\footnote{A-mode words are the most fine-grained units which are equivalent to the granularity of UniDic~\cite{maekawa2010design}.} words from the Sudachi-core dictionary.

The embedding $\mathbf{v}_w$ corresponding to word $w$ was calculated using the encoding modules of the PLMs.
Specifically, each word was converted into a sequence of subwords using the PLM tokenizers, and then sequences accompanied by special tokens such as \verb|[CLS]| and \verb|</s>| were encoded by the PLMs.
We utilized the output corresponding to a particular input token as the word-level embedding $\mathbf{v}_w$, which was used as an input for the task-specific MLP.
For example, we used the output corresponding to \verb|[CLS]| as the word-level embedding for masked language models such as BERT-base \footnote{We did not use pooling to obtain word-level embeddings according to \newcite{sun2023characters}, which reported that CLS outputs are superior to pooling outputs with Transformer.}.
For single- directional language models such as LLaMA2, we used the output corresponding to the final token.
Because the PLMs considered in this study have different special tokens and model architectures, we selected different tokens whose outputs were used as word-level embeddings.
Table \ref{tbl:input_example} summarizes the outputs for which the tokens were used as word-level embeddings.

Embeddings learned by non-Transformer architectures (i.e., Word2Vec, FastText, and GloVe) are out of focus for word-level examinations because these architectures do not have language models for encoding a sequence of tokens.

\begin{table}[t]
\centering
\small
\begin{tabular}{l|l}
\hline
Model             &  Tokenized Subword Inputs                              \\\hline
BERT-base-(un)cased &\u{[CLS]} wu \#\#gs [SEP] \\
T5-base           &\_Wu g s \u{\textless /s\textgreater} \\
LLaMA2-7B(-chat)        &\textless s\textgreater \_W \u{ugs}                            \\
CANINE            &\u{\textbackslash ue000} W u g s \textbackslash ue001                              \\
rinna-medium      &   W ug s \u{\textless /s\textgreater} \\
Swallow-7B(-Instruct)      & \textless s\textgreater \_W \u{ugs} \\ \hline
\end{tabular}
\caption{
    \label{tbl:input_example}
    Tokenization examples of an unknown word ``Wugs'' with tokenizers provided by each PLM.
    After being encoded by the PLMs, the outputs corresponding to the underlined tokens were used as word-level embeddings $\mathbf{v}_w$.
}
\end{table}

\begin{table*}[t!]
\centering
\small
\begin{tabular}{l|cc:c:cc|cc:c:cc}
\hline
                  & \multicolumn{5}{c|}{Subword-level}                                                                                            & \multicolumn{5}{c}{Word-level}                                                                                               \\ \hline
                  & \multicolumn{2}{c:}{Length} & \multirow{2}{*}{\begin{tabular}[c]{@{}r@{}}Sub-\\string\end{tabular}} & \multicolumn{2}{c|}{Constitution} & \multicolumn{2}{c:}{Length} & \multirow{2}{*}{\begin{tabular}[c]{@{}r@{}}Sub-\\string\end{tabular}} & \multicolumn{2}{c}{Constitution} \\
                  & Reg.         & Cls.         &                                                                   & Fwd.         & Bwd.         & Reg.         & Cls.         &                                                                   & Fwd.         & Bwd.        \\
                  & (MSE)      & (F1\%)     & (F1\%)                                                            & (ACC)       & (ACC)       & (MSE)      & (F1\%)     & (F1\%)                                                            & (ACC)       & (ACC)      \\ \hline
\textit{English}  &&&&&&&&&& \\
Word2Vec          & 9.48       & 26.85      & 99.68                                                             & 21.48       & 18.93       & -          & -          & -                                                                 & -           & -          \\
FastText          & 2.18       & 30.23      & 99.62                                                             & 37.67       & 25.91       & -          & -          & -                                                                 & -           & -          \\
GloVe             & 4.43       & 19.06      & 98.25                                                             & 19.38       & 14.36       & -          & -          & -                                                                 & -           & -          \\ \hdashline
BERT-base-uncased & 1.52       & 42.68      & 99.00                                                             & 38.80       & 22.52       & 1.16       & 41.37      & 98.58                                                             & 38.77       & 32.81      \\
BERT-base-cased   & 1.48       & 41.69      & 99.63                                                             & 39.54       & 26.35       & 3.79       & 39.14      & 98.59                                                             & 37.57       & 34.30      \\
T5-base           & 1.61       & 27.90      & 99.83                                                             & 28.89       & 13.05       & 0.75       & 52.30      & 97.56                                                             & 48.10       & 42.37      \\
LLaMA2-7B         & 0.35       & 75.37      & 99.93                                                             & 51.39       & 41.93       & 0.35       & 52.45      & 98.21                                                             & 55.15       & 47.37      \\
LLaMA2-7B-chat    & 0.54       & 60.99      & 99.93                                                             & 51.68       & 40.60       & 1.00       & 46.27      & 98.13                                                             & 54.85       & 47.87      \\ \hdashline
CANINE            & -          & -          & -                                                                 & -           & -           & 0.38       & 88.81      & 99.91                                                             & 83.53       & 77.27      \\ \hline
\textit{Japanese}  &&&&&&&&&& \\
rinna-medium            & 0.34      & 79.44     & 99.85                                                            & 21.49      & 19.79           & 0.30       & 68.73      & 99.18                                                             & 54.90        & 51.83      \\ 
Swallow-7B            & 0.21      & 83.40     & 99.94                                                             & 49.85       & 38.94       & 0.72     & 54.15     & 98.37         & 50.93     & 47.25      \\ 
Swallow-7B-Instruct            & 0.20 & 84.62      & 99.93     & 50.19                                                             & 38.96       & 0.75      & 51.43     & 98.44       &  48.62     & 45.81     \\ \hline
\end{tabular}
\caption{
    Experimental results of the three examinations introduced in \S \ref{sec:length-explanation}, \ref{sec:substring-explanation}, and \ref{sec:character-explanation}. We used mean squared error (MSE), weighted F1-score (F1), and accuracy (ACC) as evaluation metrics.
    Reg. and Cls. indicate the evaluations of length prediction using the regression and classification methods, respectively.
    Fwd. and Bwd. indicate $N$th character prediction from the head and tail in the forward and backward directions, respectively.
}
\label{tbl:main_result}
\end{table*}

\section{Knowledge of Token Length}
\label{sec:length-explanation}
\subsection{Examination Method}
If a PLM has acquired surface knowledge, then it should be possible to reconstruct the length of the surface from the learned embeddings.
For example, it should be possible to determine that the number of characters is four from the embedding corresponding to the word ``word.''
In this examination, we trained a regression model $f_{\mathrm{length}}$ that predicts the word length $l_{w}$ from the word embedding $\mathbf{v}_w$ (see the left side  of Figure \ref{fig:outline}).
\begin{align}
    l_{w} = f_{\mathrm{length}}(\mathbf{v}_w). \label{eq:length}
\end{align}
Here, $f_{\mathrm{length}}(\cdot)$ is an MLP\footnote{For all MLPs used in this study, the number of layers was three and the activation function was ReLU. Additionally, the embeddings of subwords/words were frozen during the training of MLPs. The size of hidden parameters was 2,096.}. We trained the MLP parameters to minimize the mean squared error (MSE) between the predicted value and the true number of characters.

For the experiment with subword-level inputs, we used the subword embedding $\mathbf{v}_s$ instead of $\mathbf{v}_w$ in Eq. \eqref{eq:length}.
The tokenizers of some PLMs append a special prefix to subwords to indicate that the subword begins from an intermediate of a word (e.g., \verb|##| in BertTokenizer) or from the head of a word (e.g., \verb|U+2581| in SentencePiece~\cite{kudo2018sentencepiece}).
Because this study focused on the surface information of the original text, we ignored such special prefixes when counting the number of characters.
For example, the length of a subword ``\#\#string'' is six.

\subsection{Results}
\label{sec:length-result}

In Table \ref{tbl:main_result}, the columns labeled ``Length'' present the performance for predicting the lengths of subwords and words using the trained MLP, which was evaluated using two metrics.
The columns labeled ``Reg.'' present the MSE scores of the MLP evaluated as a regression model.
Because the values of MSE are unintuitive, we also present weighted F1-scores, which were calculated by rounding the predicted float values to integers and then evaluating the rounded values by considering the MLP as a classification model (``Cls.'' in the table).

The experimental results highlight the difficulty of training an MLP to extract token length information from classical embedding methods such as Word2Vec.
This implies that the token length information of a surface is not acquired properly when using classical embedding methods.

In the subword-level experiments, we were able to train MLPs that appropriately predicted the token lengths when using modern PLMs with larger numbers of parameters (i.e., LLaMA2-7B) than BERT and T5.
This result demonstrates that a large number of parameters and large training corpora contribute to the acquisition of surface information in subword-level embeddings\footnote{As mentioned in \S \ref{sec:target}, the differences in scores may stem from differences in vocabularies and corpora.}.

\begin{figure}[t]
    \centering
    \includegraphics[width=7.0cm]{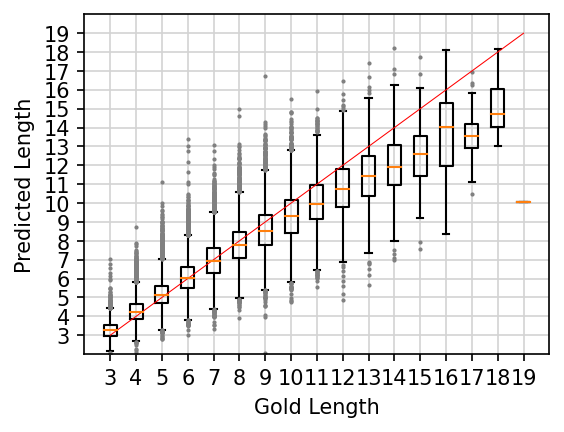}
    \caption{
        Comparison of predicted lengths and true lengths of word-level inputs in the BERT-base-cased.
        The red line indicates correct predictions.
    }
    \label{fig:bert-length-word-box}
\end{figure}

In the word-level experiments, larger models were  not clearly superior to smaller models.
These results imply that the encoders of PLMs are not capable of handling the length information stored in subword-level embeddings.
Figure \ref{fig:bert-length-word-box} presents a box plot comparing the predicted and  actual lengths of inputs from the word-level experiment with \verb|BERT-base-cased|, which has the lowest scores in Table \ref{tbl:main_result}.
One can see that the encoded word embeddings retained almost accurate length relationships, even though the exact word lengths could not be reconstructed from the embeddings.
Additionally, one can see that a longer surface makes it more difficult to predict a correct length. This may be because most of the input tokens used to train the MLPs were relatively short (approximately three to seven characters).

Although the results of the Japanese PLMs were similar to those of the English PLMs, higher weighted F1-scores were achieved by the Japanese models.
This is likely because Japanese words and subwords are typically shorter than English tokens, and it is easier to predict the length of a shorter surface.
The fact that CANINE scored the highest in the word-level experiments demonstrates that accurate length prediction can be achieved if a model can access surface information such as characters.


\section{Knowledge of Substrings}
\label{sec:substring-explanation}
\subsection{Examination Method}
If a PLM has acquired surface knowledge, then it should be possible to reconstruct substring information from learned embeddings.
For example, it should be possible to determine that ``ord'' is contained in ``word,'' whereas ``old'' is not, purely based on embeddings.
For this examination, we trained a binary classification model $f_{\mathrm{substring}}$ to predict the probability $p(t|w)$ that a string $t \in W$ is a substring of word $w \in W$ based on the corresponding embeddings $\mathbf{v}_t$ and $\mathbf{v}_w$.
\begin{align}
   p(t | w) = \sigma(f_{\mathrm{substring}}(\mathbf{v}_w \oplus \mathbf{v}_t))
\end{align}
Here, $\sigma$ is the sigmoid function and $f_{\mathrm{substring}}(\cdot)$ is an MLP.
$\oplus$ indicates vector concatenation.
$\mathbf{v}_t$ is calculated in the same manner as $\mathbf{v}_w$ using the PLMs (\S \ref{sec:gran_word}).
Among all possible pairs $\{t, w\}$, where the length of $t$ was shorter than $w$, the pairs in which $t$ was a substring of $w$ were considered as positive examples.
We randomly selected the same number of negative pairs for training.
The model was trained to minimize cross-entropy loss.
We evaluated the trained model using all possible pairs of substring candidates and words in the testing splits.
An output with $0.5 < p(t | w)$ was recognized as a positive prediction ($t$ is a substring of $w$) and the weighted F1-score was used to measure prediction performance.

For the experiment using subword-level embeddings, we used $\mathbf{v}_s$ instead of $\mathbf{v}_w$.
Additionally, we collected positive and negative samples from the vocabulary of PLMs $S$ instead of $W$.

\subsection{Results}
\label{sec:substring-result}
The columns labelled ``Substring'' in Table \ref{tbl:main_result} present the results of the experiment on substring information.
The experimental results reveal that all compared models could classify substrings almost perfectly using either subwords or word embeddings (97.5\% < weighted F1-scores).
This implies that pretrained embeddings contain information regarding whether a given string is a substring  of longer strings.
Considering that classical embedding methods such as Word2Vec contribute to almost perfect classification, it can be concluded that it is relatively easy to obtain substring information, even when using larger models.
This insight is consistent with the results of previous research~\cite{kaushal2022tokens}.

Although the results of the word-level experiments were slightly worse than those of the subword-level experiments, the trained MLPs still recognized the substrings in almost all cases.
One can observe similar trends in the results in English and Japanese.
Similar to the length experiments (\S \ref{sec:length-result}), CANINE with character-level inputs achieved the best results on the word-level experiments, thereby demonstrating the effectiveness of utilizing characters to capture surface information.

\section{Knowledge of Token Constitution}
\label{sec:character-explanation}
\subsection{Examination Method}
If PLMs have acquired knowledge regarding a text surface, it should be possible to recover the token constitution (i.e., characters comprising words) using only the learned embeddings.
For example, it should be possible to determine that ``word'' is composed of ``w + \#\#o + \#\#r + \#\#d'' based solely on the learned embeddings.
In contrast to the experiment in \S \ref{sec:substring-explanation}, this examination focused on knowledge regarding embeddings in terms of which characters are located in which positions.
To examine the obtained knowledge regarding token constitution in the embeddings, we trained a classification model that predicted the $N$-th character in the word $w$ from the embedding $\mathrm{v}_w$.
Specifically, the model calculated the probability $p(c | w, N)$, where the $N$th character of word $w$ is $c$, as follows:
\begin{align}
    \mathbf{h}^{\mathrm{N}}_w &= f_{N}(\mathbf{v}_w), \label{eq:composit_char_encode} \\ 
    p(c|w, N) &= \frac{\exp({\mathbf{h}^{\mathrm{N}}_{w}}^{\top}\mathbf{v}_{c})}{\sum_{t \in S_c}{\exp({\mathbf{h}^{\mathrm{N}}_{w}}^{\top}\mathbf{v}_{t})}}, \label{eq:composit_char_decode}
\end{align}
where $S_c \subset S$ is a subset of subwords with a token length of one\footnote{The token length count ignores special prefixes in the same manner as described in \S \ref{sec:length-explanation}.} (i.e., characters) in the vocabulary of the PLMs.
$\mathbf{v}_c$ and $\mathbf{v}_t$ are the subword embeddings corresponding to single-length subwords $c \in S_c$ and $t \in S_c$, respectively.
$f_{N}(\cdot)$ is an MLP that predicts the $N$th character, and we trained different MLPs for each position $N$.
For example, $f_{N=1}(\cdot)$ is an MLP that  predicts the first character of the surface corresponding to the input embedding.
This examination targeted an experimental setting with $N=\{1, ..., 10\}$ (i.e., from the first to the tenth character of the surface).
We conducted two types of experiments in this examination, which involved counting $N$ in the forward and backward directions from the head and tail of a word.
We trained the classifiers to minimize cross-entropy loss.

For the subword-level experiment, we used $\mathbf{v}_s$ instead of $\mathbf{v}_w$ in Eq. \eqref{eq:composit_char_encode}.
In both experiments for the word and subword levels, we used the subword-level embeddings of the PLMs for $\mathbf{v}_c$ and $\mathbf{v}_t$ in Eq. \eqref{eq:composit_char_decode}.
Similar to the other experiments, we ignored special prefixes in subwords.
For example, the first character in ``\#\#string'' is considered to be ``s.”

\begin{figure}[t]
    \centering
    \includegraphics[width=7.5cm]{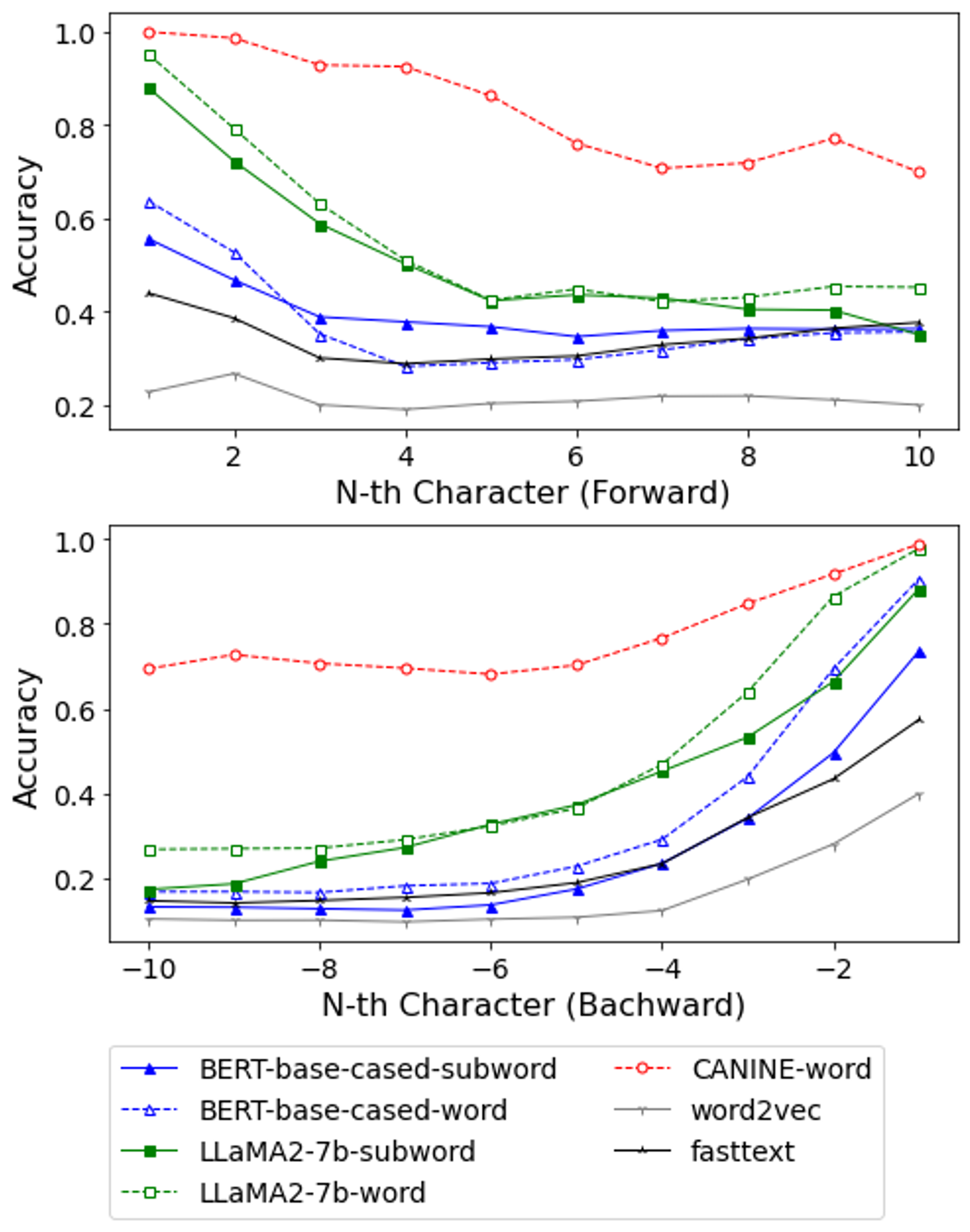}
    \caption{
        Prediction accuracy of the $N$th character in forward (top) and backward (bottom) experimental settings.
        $N=-2$ indicates the prediction accuracy for the second character counted in the backward direction from the word tail.
    }
    \label{fig:nth-character}
\end{figure}

\subsection{Results}
In Table \ref{tbl:main_result}, the columns labeled ``Constitution'' present the result of the experiments described in this section. 
The columns labeled ``Fwd.'' and ``Bwd.'' list the results of the experiments for predicting the $N$th characters counted in the forward and backward directions, respectively.

One can see that the trained MLPs could not correctly predict the $N$th characters in the subwords and words.
Even recent LLMs such as LLaMA2 could only predict correct characters with approximately 50\% accuracy.
Considering the excellent results for substring knowledge described in \S \ref{sec:substring-explanation}, the results described in this section imply that pretrained embeddings store information regarding which characters are included in text but not information regarding where characters are located within strings.

Overall, the performance of the embeddings learned by the recent LLMs like LLaMA2 was higher than that of the other embeddings.
This result indicates that the number of trainable parameters and training corpora are important factors determining whether a model is capable of obtaining knowledge regarding token constitution.

The results of the experiments in the forward setting were better than those in the backward setting for all compared models.
This indicates that the trained MLPs could predict characters more accurately when they appeared near the beginning of an input word or subword.

We observed the same trends in the experimental results for the English and Japanese models, even though Japanese tokens are comprised of smaller numbers of characters compared with English tokens.
These results indicate that limited knowledge regarding token constitution in pretrained embeddings is a common problem across two languages. 

Similar to the results described in \S \ref{sec:length-result} and \S \ref{sec:substring-result}, the results of CANINE were better than those of the other subword-level architectures. 
This indicates that the prediction task of token constitution is valid for examining the surface knowledge of PLMs because this task can be completed if models can access raw surface information such as characters. 


Figure \ref{fig:nth-character} presents the variations in accuracy with the position $N$ of the predicted characters in the text.
For the sake of visibility, we selected two types of classical embedding methods (i.e., Word2Vec and FastText), four variations of subword-level PLMs (i.e., BERT and LLaMA2), and the character-level PLM CANINE.
One can see that the trained MLPs can correctly predict the characters for small values of $N$ such as $N={1, 2, 3}$ in the forward setting (top of the figure) and $N={-1, -2, -3}$ in the backward setting (bottom of the figure).
However, for positions far from the beginning or end of a token, accuracy degrades to less 50\%.
Because the tokens associated with smaller $N$ settings include shorter words and subwords, it is not difficult to predict the characters located in specific positions.
Overall, it can be concluded that the trained MLPs had difficulty extracting surface information for long tokens and positions far from the beginning or end of a token.

The scores for the positions before $N < -5$ in the backward setting were lower than those for similarly distant positions in the  forward setting ($5 < N$).
This result implies that it is difficult to store exact character information for characters located near the tail of the subwords or words in pretrained embeddings.
However, the prediction scores of the final two characters ($N={-1, -2}$) were slightly higher than those of the first two characters ($N={1, 2}$).
This is likely because the suffixes of English words are determined by their parts of speech (e.g., ``-ly'' endings for adverbs).


\section{Evaluation with Generation}
\begin{table}[t]
\centering
\small
\begin{tabular}{lr}
\hline
Model             & Performance (F1\%) \\ \hline
BERT-base-uncased & 9.97        \\
BERT-base-cased   & 2.96        \\
LLaMA2-7B         & 3.54        \\
LLaMA2-7B-chat    & 21.08       \\ \hline
\end{tabular}
\caption{
    Prediction performance of word length using the generation method in a zero-shot manner.    \label{tbl:generation_eval}
}
\end{table}
This section discusses the surface information acquired by PLMs from the perspective of generated results.
The examinations introduced in the previous sections (see \S\ref{sec:length-explanation}, \ref{sec:substring-explanation}, \ref{sec:character-explanation}) were limited to the analysis of the intra-knowledge contained in PLMs (i.e., embeddings and encoder parameters of PLMs).
In this section, we examine whether PLMs can generate knowledge regarding surface information, specifically length knowledge.

We conducted an experiment using two English PLM families, namely BERT and LLaMA2.
We observed the generation scores of subwords corresponding to the numbers 1 to 20 based on the following prompt: ``The number of characters in \{WORD\} is...''\footnote{We selected the best-performing prompt among various candidates. For the BERT inputs, we added [MASK] to the end of the prompt and calculated the prediction score corresponding to the mask token.}.
In other words, we tested the zero-shot performance of BERT and LLaMA2 to evaluate knowledge regarding surface length.
For the placeholder ``\{WORD\}'' in the prompt, we utilized a word list consisting of the most frequently appearing 1,000 words provided by Wiktionary. 
We considered the tokens of the number with the highest generated score as the prediction for the length of each input word.

Table \ref{tbl:generation_eval} presents the performance of the model outputs, which was evaluated in terms of classification using weighted F1-scores, similar to the experiments described in \S \ref{sec:length-result}.
The best performance was obtained by LLaMA2-7b-chat, which was trained using the instruction data .
There are large gaps between the scores reported here and those obtained in \S \ref{sec:length-result} (the ``Length-Cls.'' column for the word-level experiments in Table \ref{tbl:main_result}).
For example, the highest score achieved by LLaMA2-7B-chat was 21.08 in this experiment but 46.27 in the experiment using embeddings, as reported in Table \ref{tbl:main_result}.
We cannot directly compare these two scores because we used different word list sizes. However, the experimental setting here should be easier than that described in \S \ref{sec:length-result} because the 1,000 most frequently appearing words used in this evaluation should be frequent even in the pretraining of the PLMs and their corresponding embeddings should be sufficiently robust.

As shown in Figure \ref{fig:bert-length-word-box}, knowledge regarding length information appears to be stored in word embeddings, particularly for frequently  appearing shorter words.
The observed performance gaps between the examination with the embeddings and that with generated outputs imply that PLMs cannot appropriately handle knowledge regarding surface information (i.e., length information in this setting), although such knowledge is contained in the parameters of PLMs.
In other words, there are likely additional bottlenecks on the decoder side such as shortcomings in PLM numeracy~\cite{spithourakis2018numeracy} and the inability of PLMs to handle positional information~\cite{dufter2022position}.

\section{Conclusion}
This study analyzed the knowledge regarding surface information acquired by PLMs.
We examined the knowledge obtained by six English PLMs and three Japanese PLMs in addition to three classical embedding methods.
The knowledge of PLMs was examined by training MLPs to extract surface information from subword- and word-level embeddings, namely the lengths, substrings, and token constitution of input texts.

The experimental results revealed that recent PLMs such as LLaMA2 and Swallow can acquire knowledge regarding lengths and substrings.
However, the results indicate that PLMs do not sufficiently acquire order information such as which character appears at which position.
These results imply that surface knowledge is not perfectly obtained by PLMs. This shortcoming will reduce performance on NLP tasks requiring models to handle surface information.
In addition to a lack of knowledge regarding embedding parameters, we identified bottlenecks on the decoder side in terms of utilizing acquired knowledge for generation.

In this study, we highlighted the insufficiency of knowledge regarding surface information learned by PLMs.
We believe that continuing to observe and analyze the characteristics of knowledge regarding surface information acquired by PLMs will help improve NLP performance with both PLMs and LLMs.
We will continue to examine surface knowledge in multiple languages and develop benchmarks for evaluating basic LLM abilities.


\section*{Limitation}
This study was limited to two languages (English and Japanese) with different writing system characteristics.
Experiments with languages with right-to-left writing systems such as Arabic and other languages with low-resource corpora could lead to different conclusions.
Additionally, our experiments focused on LLMs with relatively small numbers of parameters (up to 7B) in consideration of limited computational resources.
Although this shortcoming could be resolved by using larger numbers of parameters, current state-of-the-art models such ChatGPT seem to be incapable of handing surface information, as shown in Figure \ref{fgr:intro_example}.
We excluded models which are accessible only via API such as ChatGPT from our experiments because we could not guarantee the reproducibility of output results and because concrete architectures and parameters are not publicly available.
All experiments were conducted on NIVIDIA RTX A6000 and each training of the examination used in this paper was less than six hours.

\section*{Ethic Statement}
All models and data used in this study,  excluding the Word2Vec and FastText models, are publicly available.
The Word2Vec and FastText models can be reproduced using publicly available corpora.
The analysis presented in this paper may be affected by the characteristics of different languages (i.e., English and Japanese).
The experimental results should not be used to discuss  language superiority.

\section*{Acknowledgement}
These research results were obtained from the commissioned research (No.22501) by National Institute of Information and Communications Technology (NICT) , Japan.
This work was also partially supported by JST, ACT-X Grant Number JPMJAX21AM, Japan.

\bibliography{custom}


\end{document}